\documentclass[default,iicol]{sn-jnl}

\usepackage{graphicx}%
\usepackage{float}%
\usepackage{multirow}%
\usepackage{amsmath,amssymb,amsfonts}%
\usepackage{amsthm}%
\usepackage{mathrsfs}%
\usepackage[title]{appendix}%
\usepackage{xcolor}%
\usepackage{textcomp}%
\usepackage{manyfoot}%
\usepackage{booktabs}%
\usepackage{algorithm}%
\usepackage{algorithmicx}%
\usepackage{algpseudocode}%
\usepackage{listings}%
\usepackage{mhchem}
\usepackage{soul}

\begin{document}

\title[Article Title]{Fast and scalable retrosynthetic planning with a transformer neural network and speculative beam search}


\author*[1]{\fnm{Mikhail} \sur{Andronov}}\email{mikhail.andronov@idsia.ch}\equalcont{These authors contributed equally to this work.}

\author[5]{\fnm{Natalia} \sur{Andronova}}\equalcont{These authors contributed equally to this work.}

\author[4]{\fnm{Jürgen} \sur{Schmidhuber}}

\author[1,3]{\fnm{Michael} \sur{Wand}}

\author[2]{\fnm{Djork-Arné} \sur{Clevert}}

\affil*[1]{\orgname{SUPSI, IDSIA}, \orgaddress{\city{Lugano}, \country{Switzerland}}}

\affil[2]{\orgdiv{Machine Learning Research}, \orgname{Pfizer Research and Development}, \orgaddress{\city{Berlin}, \country{Germany}}}

\affil[3]{\orgdiv{Institute for Digital Technologies for Personalized Healthcare}, \orgname{SUPSI}, \orgaddress{\city{Lugano}, \country{Switzerland}}}

\affil[4]{\orgdiv{Center for Generative AI}, \orgname{KAUST}, \orgaddress{\city{Thuwal}, \country{Saudi Arabia}}}

\affil[5]{Independent researcher}


\abstract{AI-based computer-aided synthesis planning (CASP) systems are in demand as components of AI-driven drug discovery workflows. However, the high latency of such CASP systems limits their utility for high-throughput synthesizability screening in de novo drug design. We propose a method for accelerating multi-step synthesis planning systems that rely on SMILES-to-SMILES transformers as single-step retrosynthesis models. Our approach reduces the latency of SMILES-to-SMILES transformers powering multi-step synthesis planning in AiZynthFinder through speculative beam search combined with a scalable drafting strategy called Medusa. Replacing standard beam search with our approach allows the CASP system to solve 26\% to 86\% more molecules under the same time constraints of several seconds. Our method brings AI-based CASP systems closer to meeting the strict latency requirements of high-throughput synthesizability screening and improving general user experience.
}

\maketitle

\section{Introduction}\label{sec1}
Modern pharmaceutical industry is heavily betting on Artificial Intelligence (AI) technologies in an effort to reduce the enormous time and money costs of the development of new drugs \cite{Vijayan2022}. So far, AI tools have been most impactful in the preclinical stage of drug discovery, becoming an integral part of the traditional Design-Make-Test-Analyze (DMTA) cycle. Numerous solutions for \textit{de novo} drug design offer an arbitrary amount of AI-generated molecular structures for the first stage of the DMTA cycle, already helping to identify potential drug candidates \cite{Ivanenkov2023}. Naturally, the initial generated molecules must undergo extensive filtering, a part of which is filtering for synthesizability. Synthesizability, i.e., the existence of a valid synthesis route from a given molecule to the available building blocks, may depend on factors such as route length, yield, cost, the available stock of building blocks, the guidelines for allowed reaction types, etc \cite{Stanley2023}. While there are methods for synthesizability assessment based on molecular complexity scores \cite{Ertl2009} or classification of molecular structures \cite{Liu2022,Hassen2025}, the most precise and flexible way of assessing a molecule's synthesizability is constructing its complete retrosynthetic tree with a Computer-Aided Synthesis Planning (CASP) System. 

Like other areas of drug discovery, synthesis planning is also being transformed with AI, and AI-powered CASP systems are now in demand. Open-source AI-based CASP Systems such as AiZynthFinder \cite{Saigiridharan2024,Genheden2022}, ASKCOS \cite{Tu2025}, SynPlanner \cite{Akhmetshin2025}, and Syntheseus \cite{Maziarz2025} combine a single-step retrosynthesis model and a planning algorithm (e.g., MCTS \cite{Segler2018} or A* \cite{Chen2020}), implementing the design proposed by Segler et al. \cite{Segler2018}

A key challenge limiting the integration of AI-based CASP systems into the DMTA cycle is in the harsh latency requirements that a CASP tool must meet in order to keep up with the flood of structures output from \textit{de novo} generators. Current AI CASP systems are not fast enough for applications in the high-throughput setting, taking seconds to hours to solve a molecule \cite{Torren2024,Hassen2022mind}. Therefore, AI CASP systems will greatly benefit from their inference acceleration.

The single-step retrosynthesis models that enable state-of-the-art accuracy are template-free models based on a general sequence-modeling neural network architecture called the transformer \cite{Vaswani2017,Schmidhuber1992,Schlag2021}. 
Typically, the transformer-based single-step retrosynthesis model "translates" a query product SMILES into a set of candidate precursor SMILES using beam search in inference \cite{Schwaller2019,Tetko2020,Irwin2022}. Since transformers also serve as the backbone for most Large Language Models (LLMs) \cite{Brown2020}, SMILES-to-SMILES transformers as single-step models provide unique opportunities for latency optimization inspired by advances in LLM inference acceleration.

Recently, we proposed speculative beam search (SBS) \cite{Andronov2025}, an extension of speculative decoding that allows accelerated generation of multiple target sequences per query sequence, and demonstrated how it improves the latency of a SMILES-to-SMILES transformer for single-step retrosynthesis.

In the present work, we demonstrate the acceleration of multi-step retrosynthesis that relies on a SMILES-to-SMILES transformer as a single-step model by means of speculative beam search using AiZynthFinder. We combine our SBS with a state-of-the-art drafting approach called Medusa and achieve significant speed gains in multi-step retrosynthesis on Caspyrus10k \cite{Bequignon2023} under tight time constraints.

\section{Methods}
\subsection{Speculative Decoding}
Speculative decoding is a method of reducing the generation latency of autoregressive transformer models. It was originally introduced in the field of Large Language Models (LLM) research, where the inference speed of the models is a critical issue. Autoregressive transformers are the foundation of both LLMs and template-free SMILES-to-SMILES synthesis prediction, they generate one text token per run, requiring multiple sequential runs to complete a text or a SMILES string. Each model run (forward pass) involves substantial computational overhead and processing time. Speculative decoding accelerates generation by reducing the number of model calls without sacrificing accuracy. It does so by accepting or rejecting entire subsequences that serve as guesses for potential sequence continuations, where these guesses, called "drafts", may come from arbitrary sources. For example, they can be generated by a dedicated draft model or assembled based on heuristics.

The transformer decoder accepts a sequence of $N$ tokens as input and predicts the next token for each position. In standard autoregressive generation, we discard all predictions except the last one, append it to the input sequence, and run the transformer again. However, in speculative decoding, we first concatenate the input sequence with a draft sequence of $M$ tokens to leverage predictions for multiple positions simultaneously. If the prediction for the last input token matches the first token in the draft sequence, we accept the first draft token and check the prediction for the next position. We repeat this process until either a predicted token differs from the corresponding draft token, or we reach the end of the draft sequence. This approach generates between 1 token in the worst case and $M+1$ tokens in the best case per forward pass of the transformer.

One measure of the success of speculative decoding is the \textit{acceptance rate}. It is the probability of accepting a token from the draft \cite{Leviathan2023}. The empirical mean acceptance rate on the test set is the proportion of accepted speculative tokens to all speculative tokens.

\subsection{Speculative Beam Search}
The limitation of the basic speculative decoding is that it only supports generating one output sequence per input sequence. While sufficient for text generation in most cases, it is a major obstacle limiting the potential of speculative decoding for the acceleration of template-free reaction prediction and single-step retrosynthesis models. When working as a component of a CASP system, a SMILES-to-SMILES transformer must produce multiple candidate predictions for every query, typically through beam search. We recently developed a method called "speculative beam search (SBS)" \cite{Andronov2025} in an attempt to introduce speculative decoding to CASP. SBS achieves up to 3X acceleration of Molecular Transformer \cite{Schwaller2019} inference for reaction prediction and single-step retrosynthesis. The core idea of SBS is an extra step before appending the accepted draft tokens to the growing sequence. After deciding on the accepted tokens, we determine the top-K most probable next tokens for every accepted token through the forward pass of the model. With that, we obtain a set of candidate subsequences of different lengths, which we then sort by probabilities and extract top-K most probable continuations to use as beams. Both shorter and longer sequences may be the most probable. The drafts in SBS come from a heuristic drafting scheme in which multiple fragments of the query SMILES are used as drafts.
\bmhead{Heuristic drafting for SMILES generation}
SMILES-to-SMILES generation is one task that is remarkably well-compatible with speculative decoding. In chemical reactions, only some of the reactant atoms typically change their connectivity, while large fragments of the reactants remain unchanged and appear the same in the products. Therefore, instead of constructing the target SMILES token-by-token, the transformer can quickly assemble it out of fragments of the query SMILES if they are presented as draft sequences. Extracting multiple fragments of a fixed length from a query sequence, trying them all as drafts at every generation step and choosing the draft with the most accepted tokens is the essence of the heuristic drafting scheme for the SBS algorithm \cite{Andronov2025}.

\begin{figure*}
\vskip 0.2in
\begin{center}
\centerline{\includegraphics[width=1.0\textwidth]{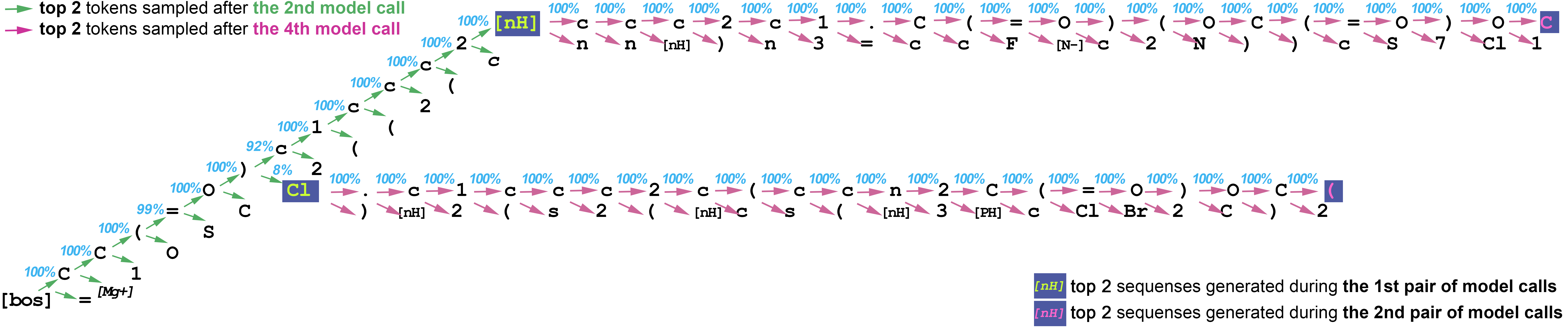}}
\caption{An example of two first cycles of the sampling of candidate sequence trees in our Medusa speculative beam search with beam size 2. Each cycle takes 2 model calls. Here, we select the two best candidates at each cycle. The first model call produces 'CC(=O)c1ccc2ccccc2.C' as the draft of 20 tokens for 'CC(=O)c1ccc2c(ccn2C(=O)OC(C)(C)C)c1' as the encoder input and '[bos]' as the decoder input. Then this draft is concatenated to the decoder input and another model call produces probabilities. The predicted probabilities of the main model head are used to make top-p (nucleus 99.75\%) verification (12 draft tokens are accepted) and to produce candidates in top-k mode. The best sequences are \textbf{CC(=O)c1ccc2} and \textbf{CC(=O)Cl}, and they become the "beams". In the next iteration, the draft for the first "beam" is \textbf{ccc2c1.C(=O)(OC(=O)O}(all 20 tokens are accepted), and for the second one it is \textbf{.c1ccc2c(ccn2C(=O)OC}(all 20 tokens are accepted). The fourth model call generates 44 sequences overall, which all get sorted by their probabilities. The most probable sequences in the second cycle are \textbf{c1c[nH]c2ccc(C(C)=} and \textbf{c1cn(C(=O)OC(C)(C)C)c2}, and they become the generated sequences for the next iteration. In this example, after 2 pairs of model calls MSBS generates 2 sequences of lengths 35 and 28, respectively, whereas the standard beam search would have generated 2 sequences of length 4.\label{fig.1a}}
\end{center}
\end{figure*}
\begin{figure*}
\vskip 0.2in
\begin{center}
\centerline{\includegraphics[width=1.0\textwidth]{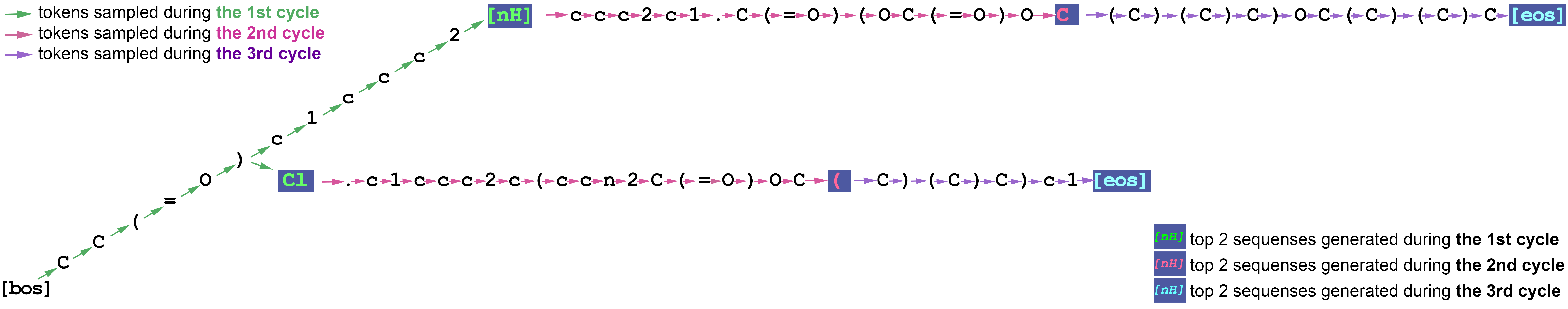}}

\label{fig:sbs_sample}
\end{center}
\vskip -0.2in
\caption{A simplified illustration of building top 2 sequences with MSBS algorithm for the source molecule \textbf{CC(=O)c1ccc2c(ccn2C(=O)OC(C)(C)C)c1}. It takes 6 model calls instead of 52 ones of classic beam search. MSBS generates \textbf{CC(=O)c1ccc2[nH]ccc2c1.C(=O)(OC(=O)OC(C)(C)C)OC(C)(C)C} (coincides with the target) and \textbf{CC(=O)Cl.c1ccc2c(ccn2C(=O)OC(C)(C)C)c1} as the result.\label{fig.1b}}
\end{figure*}

\subsection{Medusa}
While SBS achieves good inference acceleration in reaction prediction and single-step retrosynthesis, the heuristic drafting strategy presents a scalability problem \cite{Andronov2025}. To achieve a high acceptance rate, multiple drafts should be used in parallel. It increases the effective batch size as $O(BKN)$, where $B$ is the basic batch size, $K$ is the number of beams, and $N$ is the number of drafts. The latency of a forward pass of the model increases with batch size, and may quickly outweigh the benefits of the high throughput given by batching.

A recently proposed method called "Medusa" \cite{Cai2024} presents a simple solution for generating single drafts with a high acceptance rate. The fundamental idea of the method is to add extra subnetworks (decoding heads) to the transformer neural network that predict multiple tokens ahead of the next token in parallel. Instead of usual transformer logits output of shape $(B, L, V)$ a Medusa model gives $(B, L, M, V)$, where $B$ is the input batch size, $L$ is the decoder input length, $V$ is the vocabulary size, and $M$ is the number of Medusa model heads. While the main prediction head generates the next token as usual, the additional Medusa heads predict the second next token, the third next token, and so on up to the $M$-th next token. 

The tokens predicted by the additional heads are the draft sequences for the main head to verify.
The first Medusa call is used to generate a draft. In our experiments, the model has 20 heads, so the draft length is 20. We use greedy decoding to create only one draft per given input sequence to avoid inflating the effective batch size. The second Medusa call uses only the main head's output data to verify draft tokens. At least one draft token will always be approved (as it was generated by the main model head in greedy mode), and thus 2 tokens will be generated in 2 model calls in the worst case.
Of course, the worst-case scenario is still undesirable, since additional heads require additional weights in the model architecture that can make forward pass a bit slower. In our architecture, the addition of extra heads resulted in an increase in the number of weights by 7.5\%. Thus, a high acceptance rate for draft tokens is important. In the best case, the Medusa model with 20 heads (1 main head and 19 extra heads) produces 21 tokens in 2 model calls.

As a verification procedure, we use one similar to top-p sampling. We sort the predicted probabilities for every token in the vocabulary in descending order and calculate the cumulative probabilities for every token. If such cumulative probability corresponding to a given draft token is less than the nucleus parameter (we use 99.75\%) then that token is probable enough and it gets approved. Additionally, the highest probability token among all vocabulary tokens is always approved. Fig. \ref{fig.1a} and Fig. \ref{fig.1b} provide an example. 

To train the model, we choose the recipe "joint training, combined loss" from the original Medusa paper \cite{Cai2024}. To give priority to the accuracy of the main head, we divide each head's contribution to the loss function by the head's number.

When we replace the heuristic drafting for the Molecular Transformer with the Medusa approach, we observe a significant improvement in SBS scalability to larger batch sizes, expanding the potential of speculative decoding and transformer models in fast synthesis planning.

\subsection{Multi-step synthesis planning}
Since the ultimate goal of building accelerated template-free SMILES-to-SMILES prediction models is to enable fast AI-powered CASP, we evaluate our single-step retrosynthesis model as a component of a multi-step synthesis planning system. We choose AiZynthFinder \cite{Genheden2020,Saigiridharan2024} because of its straightforward support for arbitrary template-free models and to maintain continuity with prior work that benchmarked various single-step retrosynthesis models as the components of AiZynthFinder \cite{Torren2024}. We choose Retro* \cite{Chen2020} as the search algorithm for building the synthesis tree. We use only the reactant probability of the single-step model as the guiding probability in tree search building, as it was done by \citeauthor{Torren2024} \cite{Torren2024}

\subsection{Model}
We train a custom variant of the Molecular Transformer \cite{Schwaller2019}, an encoder-decoder transformer for SMILES-to-SMILES translation. Our model has six encoder and decoder layers, eight heads in multi-head attention, embedding dimensionality of 256, feedforward dimensionality of 2048, and 20 additional Medusa heads to predict the tokens from 1 to 20 positions ahead of the next token. All Medusa heads are implemented as an MLP with one hidden layer with dimensionality $20 \times 50 = 1000$, followed by a residual connection and layer normalization. The number of parameters is 17.4 million in the base transformer, and 1.3 million in the Medusa heads, resulting in 18.7 million parameters overall. We use the Adam \cite{Kingma2014} optimizer for training. We train and test our model on one NVIDIA Tesla V100 GPU with 32GB of memory.

\subsection{Data}
For training the single-step retrosynthesis models and its isolated evaluation, we employ the standard USPTO50K dataset. We apply the 20-fold R-SMILES augmentation \cite{Zhong2022} to the training subset of USPTO50K, which is beneficial for the model's accuracy. The test set comprises 5007 reactions; we do not augment it. We follow the standard atomwise tokenization procedure \cite{Schwaller2019} to tokenize SMILES.

\section{Results and Discussion}

\subsection{Single-step retrosynthesis}
We first test our transformer model in a single-step retrosynthesis setting on USPTO50K and compare three inference modes: beam search (BS), speculative beam search with a "smart" heuristic drafting variant \cite{Andronov2025} (HSBS), and speculative beam search with drafting based on tokens predicted by the Medusa heads (MSBS). We also include a separate "optimized" beam search that does not call the model to predict the pad-token after the eos-token. This optimization of the beam search does not influence the accuracy and number of model calls. However, it decreases the effective batch size to help reduce the calculation time at larger batch sizes. Since HSBS and MSBS do not call their models to produce the pad-token after the eos-token, we outline "beam search optimized" as a separate position to ensure a more accurate comparison. 

As Table \ref{single_step_time}A shows, MSBS significantly outperforms BS and HSBS at various batch sizes in terms of inference speed. HSBS outperforms BS at smaller batch sizes but suffers from scalability limitations. Due to the \textit{throughput-latency tradeoff} inherent in processing multiple draft sequences simultaneously, the heuristic drafting scheme requires careful tuning of draft number and length for optimal performance. At larger batch sizes, the computational overhead of processing multiple drafts negates the acceleration benefits, and the optimal number of drafts becomes 1, making HSBS similar to MSBS, as it also uses only one draft. At the same time, MSBS achieves a higher acceptance rate (Table \ref{single_step_time}D) through its integrated architecture, maintaining consistent acceleration even at batch size 32, which establishes MSBS as the superior acceleration approach for single-step retrosynthesis with transformers. MSBS requires fewer forward passes of the model to finish the generation (Table \ref{single_step_time}B) and boasts an acceptance rate of 91\%, leaving HSBS far behind.

In terms of accuracy and prediction validity, all three methods demonstrate nearly identical performance (Table \ref{single_step_acc}). While our speculative beam search approach does not guarantee output distributions identical to standard beam search, the practical differences prove negligible. A slightly larger difference in accuracy and SMILES validity between MSBS and HSBS stem from the marginal performance differences between model checkpoints rather than algorithmic effects: MSBS implies a custom transformer architecture and requires training a separate model, while HSBS is a drop-in replacement for beam search.

\begin{table*}[h]
\caption{Comparison between the inference algorithms for the single-step retrosynthesis model on the USPTO 50k test set (5K reactions). "Beam search optimized" means that the finished sequences in a batch are put aside and the transformer is not called to generate pad tokens after the EOS token (it reduces the average effective batch size). HSBS is speculative beam search with transformer and heuristic drafting, MSBS is speculative beam search with Medusa model. "B" stands for batch size, and "K" stands for the number of generated sequences (beam size for beam search). The number of drafts and the draft length in HSBS are individual for every B: 10 drafts of length 10 for B=1, 3 drafts of length 10 for B=4, and 1 draft of length 20 for other B. Medusa heads always generate 1 draft of length 20. The average time and the standard deviation are  estimated based on five runs.}
\label{single_step_time}
\vskip 0.15in
\begin{center}
\begin{small}
\begin{sc}
\begin{tabular}{lccccc}
\toprule
(A) Decoding wall time (K=10), min & B=1 & B=4 & B=8 & B=16 & B=32 \\
\midrule
Beam search   & 50.0 $\pm$ 3.8 & 26.9 $\pm$ 3.5 & 18.7 $\pm$ 1.2 & 14.9 $\pm$ 0.1  & 16.2 $\pm$ 0.1  \\
Beam search optimized   & 50.0 $\pm$ 2.2 & 16.2 $\pm$ 0.3 & 9.4 $\pm$ 0.2 & 7.3 $\pm$ 0.1  & 5.5 $\pm$ 0.1  \\
HSBS  & 22.7 $\pm$ 1.3 & 10.1 $\pm$ 0.2 & 7.4 $\pm$ 0.2 & 6.1 $\pm$ 0.1  & 5.2 $\pm$ 0.0 \\
MSBS  & 11.4 $\pm$ 0.4 & 4.0 $\pm$ 0.2 & 2.4 $\pm$ 0.2 & 2.1 $\pm$ 0.1 & 1.5 $\pm$ 0.1 \\
\bottomrule
\\
\midrule
(B) Model calls (K=10) & B=1 & B=4 & B=8 & B=16 & B=32 \\
\midrule
Beam search   & 295,947 & 99,030 & 54,934 & 29,941  & 16,170  \\
Beam search optimized  & 295,947 & 99,030 & 54,934 & 29,941  & 16,170  \\
HSBS  & 92,538 & 36,960 & 28,056 & 15,807  & 8,817 \\
MSBS   & 59,502 & 19,240 & 10,730 & 5,906 & 3,224 \\
\bottomrule
\\
\midrule
(C) Average effective batch size (K=10) & B=1 & B=4 & B=8 & B=16 & B=32 \\
\midrule
Beam search   & 10 & 40 & 80 & 160  & 320  \\
Beam search optimized  & 8 & 25 & 45 & 82  & 151  \\
HSBS  & 23 & 40 & 29 & 52  & 93 \\
MSBS   & 6 & 18 & 32 & 58 & 105 \\
\bottomrule
\\
\midrule
(D) Acceptance rate (K=10), \% & B=1 & B=4 & B=8 & B=16 & B=32 \\
\midrule
HSBS  & 74 & 70 & 64 & 64 & 64 \\
MSBS   & 91 & 91 & 91 & 91 & 91 \\
\bottomrule
\end{tabular}
\end{sc}
\end{small}
\end{center}
\vskip -0.1in
\end{table*}

\begin{table*}[h]
\caption{The top-N accuracy of our model in single-step retrosynthesis on USPTO 50K and the proportion of invalid SMILES in the N-th prediction with different decoding strategies: beam search, speculative beam search with heuristic drafting strategy (HSBS), speculative beam search with Medusa heads for drafting (MSBS). The difference in accuracy between all decoding methods is negligible.}
\label{single_step_acc}
\vskip 0.15in
\begin{center}
\begin{small}
\begin{sc}
\begin{tabular}{llcccc}

\multicolumn{5}{c}{} \\
\toprule
Single-step retrosynthesis & & Top-1 & Top-3 & Top-5 & Top-10 \\
\midrule
Accuracy, \% & Beam search & 52.08 & 75.16 & 82.97 & 89.08 \\
        & HSBS         & 52.08 & 75.16 & 82.07 & 89.12 \\
        & MSBS        & 54.08 & 75.99 & 82.92 & 89.23 \\
\midrule
     &    & Pred. 1 & Pred. 3 & Pred. 5 & Pred. 10 \\
\midrule

Invalid SMILES, \% & Beam search & 0.8 & 1.8 & 3.5 & 8.1 \\
        & HSBS         & 0.8 & 1.8 & 3.5 & 8.2 \\
        & MSBS & 0.4 & 1.6 & 3.1 & 9.3 \\

\bottomrule
\end{tabular}
\end{sc}
\end{small}
\end{center}
\vskip -0.1in
\end{table*}

\subsection{Multi-step retrosynthesis}
We test the performance of our single-step retrosynthesis model in the multi-step synthesis setting on the Capyrus10K dataset using AiZynthFinder for building the synthesis tree. We drew inspiration from the work of \citeauthor{Torren2024} \cite{Torren2024}; however, we adjusted the methodology to prioritize inference speed, assuming chemists would not wait for hours for computation completion. We constrain the multi-step synthesis to either solve a query molecule in several seconds or count it as unsolved. As building blocks, we use the PaRoutes \cite{Genheden2022} stock containing 13414 molecules. We generate 10 candidate precursor sets with every call of a single-step model and constrain the maximum route length to 5 and the maximum number of algorithm iterations to 35000. When the algorithm finds the first route from a query molecule to the building blocks, the algorithm stops, and the molecule is considered solved.

Table \ref{multistep_results} summarizes our multi-step retrosynthesis experiments. The results reveal that speculative beam search with Medusa heads (MSBS) consistently outperforms standard beam search (BS) across all experimental conditions, with improvements in both the number of solved molecules and computational efficiency. 

Under depth-first search with a 5-second limit, MSBS solved 2080 molecules out of 10000, which is 86\% more compared to the 1117 solved by BS. For the 1017 molecules that both methods successfully solved, MSBS required less than half the time on average (0.86s vs 1.88s). 

With the more sophisticated Retro* algorithm, MSBS maintained its advantage, solving 36\% more molecules than BS within 5 seconds (5287 vs 3890) and 26\% more within 15 seconds (6715 vs 5341). Across all conditions, MSBS consistently achieved faster average solution times while solving substantially more molecules.

Interestingly, MSBS required more algorithm iterations per commonly solved molecule than BS. This likely reflects differences in probability distributions: BS tends to concentrate probability mass on the top candidate, while MSBS produces more uniform distributions across candidates, leading to more exploratory search behavior that ultimately identifies additional solutions.

Although MSBS exhibits good scalability with batch size in batched inference, AiZynthFinder does not take advantage of it. By design, it
runs all single-step expansions with batch size 1. Therefore, in the reported multi-step synthesis experiments (Table \ref{multistep_results}), the comparison between BS and
MSBS only considers the performance at batch size 1.  Even though MSBS demonstrates clear utility even without batching support, we decided to conduct additional experiments, forcing batching into the Retro* search. We added an option to take more than one entry at a time from the priority queue in the Retro* algorithm and denoted the number of entries extracted from the queue as "beam width". In our experiments, we change this beam width from 1 to 16. In this case, the single-step model is called with a batch size of 16, so that one algorithm iteration generates 10 sets of precursors for each of the 16 given
molecules and then these precursors are added to the tree in a cycle. Although this design may not be strictly mathematically justified, it increased the percentage of solved molecules (Table \ref{retro_16}) by allowing the algorithm to generate more routes within the same time limits. The results suggest that the path to fast retrosynthesis lies in creating algorithms that rely on single-step retrosynthesis models working continuously with large batch sizes, and in developing single-step models that support batch sizes as large as possible.
\begin{table*}[h]
\caption{The comparison between two inference algorithms for the SMILES-to-SMILES transformer serving as a single-step retrosynthesis model within AiZynthFinder on Caspyrus10k under different search algorithms and time limits per molecule. BS is beam search, MSBS is speculative beam search with Medusa heads as draft source. Depth-first search (DFS) and Retro* are search algorithms that build the synthesis tree.}
\label{multistep_results}
\vskip 0.15in
\begin{center}
\begin{small}
\begin{sc}
\begin{tabular}{llcc}

\multicolumn{2}{c}{} \\
\toprule
DFS, time limit 5 seconds & & BS & MSBS \\
\midrule
Solved molecules & & 1117 & 2080 \\
Common solved molecules & & \multicolumn{2}{c}{1017} \\
Avg. time per solved molecule, s & & 2.01 & 1.85 \\
Avg. time per common solved molecule, s & & 1.88 & 0.86 \\
Avg. alg. iterations per common solved molecule & & 6.52 & 9.51 \\
\\
\toprule
Retro*, time limit 5 seconds & & BS & MSBS \\
\midrule
Solved molecules & & 3890 & 5287 \\
Common solved molecules & & \multicolumn{2}{c}{3628} \\
Avg. time per solved molecule, s & & 2.14 & 1.41 \\
Avg. time per common solved molecule, s & & 2.06 & 0.99 \\
Avg. alg. iterations per common solved molecule & & 5.51 & 7.38 \\
\\
\toprule
Retro*, time limit 15 seconds & & BS & MSBS \\
\midrule
Solved molecules & & 5341 & 6715 \\
Common solved molecules & & \multicolumn{2}{c}{5050} \\
Avg. time per solved molecule, s & & 4.25 & 2.86 \\
Avg. time per common solved molecule, s & & 4.00 & 1.84 \\
Avg. alg. iterations per common solved molecule & & 12.44 & 18.99 \\
\bottomrule
\end{tabular}
\end{sc}
\end{small}
\end{center}
\vskip -0.1in
\end{table*}

\begin{table*}[h]
\caption{The comparison between different single step inference methods within AiZynthFinder on Caspyrus10k under Retro* search algorithm with time limits 5 sec and 15 sec per molecule. BS is beam search, BS optimized doesn't call transformer to produce pad-token after eos-token, the model MSBS is speculative beam search with Medusa heads as draft source. "Bw" stands for beam width in Retro* algorithm.}
\label{retro_16}
\begin{center}
\begin{small}
\begin{sc}
\begin{tabular}{llcccc}

\multicolumn{4}{c}{} \\
\midrule
 (A)    5 sec limit \\

inference & Bw & Solved molecules, \% && Total time, h \\
\midrule
BS & 1 & 38.90 && 11.3 \\
MSBS & 1 & 52.87 && 8.7 \\
BS optimized & 16 & 53.86 && 11.7 \\

MSBS & 16 & 64.09 && 8.4 \\

\midrule
\midrule
 (B)    15 sec limit \\

inference & Bw & Solved molecules, \% && Total time, h \\
\midrule
BS & 1 & 53.41 && 26.12 \\
MSBS & 1 & 67.15 && 19.0 \\
BS optimized & 16 & 70.23 && 20.47 \\

MSBS & 16 & 75.07 && 16.12 \\

\bottomrule
\end{tabular}
\end{sc}
\end{small}
\end{center}
\vskip -0.1in
\end{table*}

\subsection{Limitations}\label{limitations}
Similarly to the paper by \citeauthor{Torren2024} \cite{Torren2024}, here we concentrate only on the speed of building the retrosynthetic graph. However, Aizynthfinder also spends a significant amount of time splitting the tree into separate routes. The more model calls are made within the time limit, the more nodes are added to the retrosynthesis tree, and the more complex the tree becomes for splitting. We alleviated this problem by allowing AiZynthFinder to extract only the successful routes in which all leaves are building blocks. This approach requires only a small fraction of the time required for the exhaustive process of extracting all retrosynthetic routes in AiZynthFinder. In case unsolved routes are required, the tree splitting algorithm can be easily optimized by extracting the most probable reactions first, instead of a random choice of unsolved reactions, and also by decreasing the maximal route number limit.     

\section{Conclusion}
We demonstrate that speculative beam search, combined with the Medusa drafting strategy, significantly accelerates multi-step retrosynthetic tree search in AI-based CASP systems. When applied to a SMILES-to-SMILES transformer serving as a single-step model in the open-source AiZynthFinder system, our approach enables the successful solution of 26\% to 86\% more molecules compared to standard beam search under realistic time constraints of 5 to 15 seconds. We show that inference acceleration techniques initially developed for large language models can be successfully adapted to chemical synthesis planning tasks, pushing the speed of AI-based CASP closer to that required in high-throughput synthesizability screening in pharmaceutical research. Our future work will focus on generalizing multi-step synthesis planning algorithms to support larger batch sizes, which should further reduce the latency in CASP systems.

\backmatter

\bmhead{Funding}
This study was partially funded by the Horizon Europe funding programme under the Marie Skłodowska-Curie Actions Doctoral Networks grant agreement “Explainable AI for Molecules - AiChemist” No. 101120466.

\bmhead{Author contributions}
MA and NA conceptualized the paper idea. NA and MA wrote the code and conducted the experiments. MA and NA wrote the manuscript with inputs from all co-authors. JS, DC and MW acquired the research funding, administered the project and provided supervision.

\bmhead{Code availability}
The code is in the GitHub repository https://github.com/Academich/faster-ml-casp

\bmhead{Availability of data and materials}
The instructions for downloading the data to replicate the results are in the GitHub repository https://github.com/Academich/faster-ml-casp

\bmhead{Competing interests}
The authors have no competing interests to declare.

\bmhead{Ethics approval  and consent to participate}
Not applicable

\bmhead{Consent for publication}
Not applicable

\bibliographystyle{chicago}
\bibliography{main}

\end{document}